\documentclass[sigconf]{aamas} 

\usepackage{dblfloatfix}
\usepackage{balance} 
\usepackage{algpseudocode}
\usepackage{algorithm}
\usepackage{caption}
\usepackage{subcaption}

\setcopyright{none}
\renewcommand\footnotetextcopyrightpermission[1]{} 

\title[DIRECT]{DIRECT: Learning from Sparse and Shifting Rewards using Discriminative Reward Co-Training }

\author{Philipp Altmann}
\affiliation{\country{LMU Munich}}
\email{philipp.altmann@ifi.lmu.de}

\author{Thomy Phan}
\affiliation{\country{LMU Munich}}

\author{Fabian Ritz}
\affiliation{\country{LMU Munich}}

\author{Thomas Gabor}
\affiliation{\country{LMU Munich}}

\author{Claudia Linnhoff-Popien}
\affiliation{\country{LMU Munich}}

\begin{abstract}
  We propose \emph{discriminative reward co-training} (DIRECT) as an extension to deep reinforcement learning algorithms.
  Building upon the concept of \emph{self-imitation learning} (SIL), we introduce an imitation buffer to store beneficial trajectories generated by the policy determined by their return.
  A discriminator network is trained concurrently to the policy to distinguish between trajectories generated by the current policy and beneficial trajectories generated by previous policies. 
  The discriminator's verdict is used to construct a reward signal for optimizing the policy.
  By interpolating prior experience, DIRECT is able to act as a surrogate, steering policy optimization towards more valuable regions of the reward landscape thus learning an optimal policy. 
  Our results show that DIRECT outperforms state-of-the-art algorithms in sparse- and shifting-reward environments being able to provide a surrogate reward to the policy and direct the optimization towards valuable areas. 
\end{abstract}

\begin{document}
\pagestyle{fancy}
\fancyhead{}
\maketitle 

\section{Introduction}

Imitation plays a central role in the human learning process \cite{human-imitation-piaget51}. Studies have shown that imitation mechanisms are deeply rooted in the human brain \cite{human-imitation-iacoboni99}.
Therefore, imitation can be considered essential for the development of human intelligence \cite{human-imitation-gabora11}. 
Turing even suggested imitation to be a central capability on which to test an artificial intelligence in the \textit{Imitation Game}~\cite{imitation-game-turing50}.

Regarding human imitation, cognitive science suggests that the ability to access memories voluntarily enabled humans to act out events that occurred in the past or might occur in the future. Those enabled them to take control over their output by voluntarily rehearsing and refining, gaining skills like reenactive play, self-reminding, and imitation learning. \cite{human-imitation-gabora11} 

This has also inspired new methods of reinforcement learning, where recent positive experience is used to guide the optimization of an agent's policy, resulting in improved exploratory properties.
Those concepts are referred to as self-imitation learning (SIL) \cite{sil-oh18}. 
Similar to human behavior, imitation is incorporated to improve the exploratory capabilities.

This is especially important for reinforcement learning in real-world scenarios where precisely constructed and smooth reward functions are not available \cite{eysenbach2021maximum}.  
In some instances the reward distribution might even shift between training and execution, a phenomena referred to as distributional shift which is crucial to tackle in order to safely deploy reinforcement learning models in the real world.
The exploration of both sparse- and shifting rewards have been shown to be challenging problems, especially for deep (reinforcement) learning algorithms \cite{safety-amodei16, simtoreal-tobin17, rajeswaran2016epopt}. 

In this paper, we address policy optimization in such challenging reward scenarios and propose discriminative reward co-training (DIRECT).
DIRECT incorporates a self-imitation buffer to store beneficial trajectories generated by former policies determined by their return. 
A discriminator, inspired by generative adversarial networks (GANs) \cite{goodfellow2020generative}, is trained to distinguish between trajectories experienced by the current policy and trajectories from this self-imitation buffer. 
If the discriminator can easily discern between the current policy and its former best deeds, we assume that the current policy does not live up to its full potential and has still room to improve---and for further training. On the other hand, if the policy's current actions easily match with its best recorded experience, then the discriminator needs to examine the experience samples more closely.
This discriminative instance thus can serve as an additional or even a surrogate reward signal aiding the policy optimization towards self-imitation of previously beneficial experiences. 
This guidance is especially helpful for sparse reward landscapes and hard exploration problems.
Overall, we provide the following contributions: 
\begin{itemize}
\item We propose discriminative reward co-training (DIRECT) to shape rewards using a self-imitation buffer $\mathcal{B}$ and the corresponding update rule, a discriminator $D$ and the corresponding training loss, and an accommodating training procedure.
\item We show the directive capabilities and gained adaptability to environmental changes of DIRECT within varying safety gridworld environments.
\item We provide benchmark comparisons against PPO, A2C, DQN and SIL and show that DIRECT achieves superior performance especially in sparse- and shifting reward settings.
\end{itemize}

\section{Background}

\paragraph{Reinforcement Learning}
We formulate our problem as \emph{Markov decision process (MDP)} \cite{puterman2014markov} $\mathcal{M} = \langle \mathcal{S}, \mathcal{A}, \mathcal{P}, \mathcal{R} \rangle$, where $\mathcal{S}$ is a (finite) set of states $s_t \in \mathcal{S}$, $\mathcal{A}$ is the (finite) set of actions $a_t$, $\mathcal{P}(s_{t+1}|s_t,a_t)$ is the transition probability function, and $\mathcal{R}(s_t,a_t)$ is the reward.

The \emph{reinforcement learning (RL)} goal is to find a policy $\pi(\cdot | s_t) : \mathcal{S} \rightarrow \mathcal{A}$, which maximizes the expected return $G(\tau)$ for all trajectories $\tau \sim \pi$
\begin{equation}\label{eq:return}
G(\tau) = \sum_{t=0}^{\infty} \gamma^{t} \cdot \mathcal{R}(s_{t}, a_{t})
\end{equation}
where $\gamma \in [0,1)$ is the discount factor.
Furthermore, let the action-value function $Q^\pi(s_t,a_t)$ and the value function $V^\pi(s_t)$ of any given action $a$ and state $s$ be the expected return following $\pi$, starting with action $a$ in state $s$ and starting in state $s$ respectively. \cite{rl_sutton18} 

\paragraph{Policy Optimization}
For the following, we consider $\pi$, $V^\pi$ and $Q^\pi$ to be deep neural networks parameterized by $\theta$ and $\phi$ respectively. 
We furthermore consider interactive training data in form of rollouts generated by the current policy acting within a finite-horizon environment, i.e. an environment that terminates episodes upon reaching a goal, termination state, or a maximum of steps, or executed actions.

To learn an optimal policy $\pi^*$ for high-dimensional state spaces, applying Q-learning updates \cite{rl_sutton18} to $Q_\phi$, referred to as Deep Q-Network (DQN), has been proposed \cite{mnih2013playing}.

Policy gradient methods directly learn the policy parameters $\theta$, where the gradient is given by $\nabla_\theta J(\pi_\theta) = \nabla_\theta \mathbb{E}_{\pi_\theta}\left[ G(\tau)\right]$ for trajectories $\tau$ generated by $\pi_\theta$, where $\theta$ can be updated via gradient ascent \cite{rl_sutton18}.  
In order to enable updates using policy rollouts of incomplete episodes, the incorporation of the value network $V$ to build the advantage $A_t^\pi=R(s_t,a_t) + \gamma \cdot V^\pi(s_{t+1}) - V^\pi(s_t)$ has been proposed and is referred to as Advantageous Actor-Critic (A2C), where $\pi_\theta$ represents the actor $V_\phi$ represents the critic \cite{a3c-mnih16}.
Further improving upon this concept, Proximal Policy Optimization (PPO) builds upon Trust Region Policy Optimization (TRPO)~\cite{trpo-schulman15} introducing a surrogate to optimize the update size in order to yield faster and more stable convergence \cite{ppo-schulman17}. 
The main PPO objective is defined as:
\begin{equation}
  \mathcal{L}_t^{\mathrm{CLIP}}\left(\theta\right) = \hat{\mathbb{E}}_t\left[\min\left( 
  r_t\left(\theta\right)\hat{A}_t, clip\left(  r_t\left(\theta\right), 1-\epsilon, 1+\epsilon \right)  \hat{A}_t 
  \right)\right],
\label{eq:ppo_l_clipped}
\end{equation}
where $ r_t$ is the action probability ratio between the current policy $\pi_\theta$ and the previous policy $\pi_{\theta_{old}}$, taken from TRPO, and $\epsilon$ is a hyperparameter \cite{ppo-schulman17}.

\paragraph{Reward Shaping}
Motivated by the assumption that a reward signal might not be optimally defined for an agent to learn the intended behavior, {Ng et al.}~\shortcite{reward-ng99} proposed to extend the given reward by supplemental metrics, like the distance to a given target, or, the estimated remaining time to reach the goal. 
This can be applied to the extent, where the environmental reward signal is completely replaced by the enhanced metric \cite{reward-ng99}.
However, in contrast to the discriminative reward we propose, the construction of reward shaping metrics often requires domain knowledge. 

\paragraph{Intrinsic Rewards}
Especially in sparse reward setting, e.g. if the agent is only rewarded at the episode end instead of receiving constant incentives, intrinsic rewards have been shown to yield improved exploitative capabilities. 
Intrinsic rewards refer to a special form of reward shaping, where the environmental reward signal is complemented by an intrinsically motivated reward supplement.
In contrast to conventional reward shaping techniques, this intrinsic motivation can mostly be constructed without the need for domain knowledge, e.g. by learning an internal model. \cite{schafer2022decoupled,pathak2017curiosity}

\paragraph{RL Safety}
Applying RL methods to real-world problems while considering safety requirements within the environment often plays an equally important role as merely maximizing the expected return. {Leike et al.}~\shortcite{safegrid-leike17} therefore developed a suite of RL environments that require an agent to act safely with regard to previously defined safety problems.
Current approaches for safe RL resemble the use of a shaped reward, where domain knowledge is incorporated to maximize safety in addition to the environmental return \cite{achiam2017constrained}. 
Distributional shifts represent a subset of said problems and a critical issue within various machine learning applications \cite{dataset_shift-candela09}. 
They refer to applying a learned model to a novel set of data containing changes not included in the training data. \cite{safety-amodei16,safegrid-leike17,dataset_shift-candela09}.
In order to produce policies that are robust to distributional shift, extending the training samples by a more diverse pool of synthetic experience has been to be shown helpful \cite{mendonca2020meta}.
Within supervised learning, adversarial methods have been shown to yield improved robustness against out-of-distribution samples \cite{liu2021stable}.

\paragraph{Adversarial Learning}
Similar to the Imitation Game~\cite{imitation-game-turing50}, adversarial methods introduce an opponent or adversary, attempting to fool the decision-making instance, typically a classifier, by injecting adversarial samples. 
Adversarial samples refer to malicious or noisy samples specifically designed to attack a certain model \cite{adversarial-kurakin16, guo2019simple}. 
As many deep learning techniques have been shown to be prone to adversarial samples, not performing as expected, failing to classify them correctly, the adversarial counterpart has been attempted to be integrated into the learning process \cite{DBLP:conf/iclr/HuangPGDA17, madry2018towards, dong2018boosting}.
In general, adversarial methods have been applied to a wide variety of learning tasks like co-evolution \cite{wang2019poet, gabor2019scenario}, or testing \cite{uesato*2018rigorous, ijcai2021p591} and recently gained further interest in data or image generation via generative adversarial networks (GANs) \cite{goodfellow2020generative}. 
The GAN objective is given by the two-player minimax game 
\begin{align}
  \min_{G} \max_{D}V\left( D,G \right) &= \mathbb{E}_{x \sim p_{data}\left(x\right) } \left[ \log D\left(x\right)\right] \\
   &+\mathbb{E}_{z \sim p_{z}\left(z\right) } \left[ \log \left(1-D\left(G\left(z\right)\right)\right)\right],
\label{eq:gan}
\end{align}
where $p_z(z)$ is a random noise variable used as an input to the Generator to learn its distribution $p_g$.
Adding labels to the training data, thus turning the procedure into a semi-supervised learning process has also been shown to benefit the learning process and quality of generated samples, as shown with conditional GANs (cGANs) \cite{cgan-mirza14}.
Applying adversarial architectures has been shown to improve a learned model's robustness against changing circumstances \cite{adversarial-szegedy13,liu2021stable}.
In addition to unsupervised tasks, like data generation, and the methodological connections to Actor-Critic methods \cite{ac-gan-pfau16}, GANs have also been shown to be closely related to inverse reinforcement learning {(IRL)} and thus being applicable to those tasks as well \cite{gan-irl-finn16}. In contrast to RL, aiming to find an optimal policy $\pi^*$, IRL aims to find a reward function that best describes a given trajectory of demonstrations and can be use to learn from experts, using imitation learning \cite{irl-ng00}. 


\section{Related Work}

Even though many recent approaches tackling the difficulties of learning from a sparse reward signal have been proposed, real-world scenarios exist where a reward signal is not available at all, or the goal is to learn from dedicated expert behavior. 
For these scenarios, imitation learning approaches have been developed. \cite{schaal1999imitation, hussein2017imitation, gail-ho16, sil-oh18}

\paragraph{Generative Adversarial Imitation Learning (GAIL)}
was proposed by {Ho et al.}~\shortcite{gail-ho16} attempting to simplify the two-step procedure required to imitate expert demonstrations using {IRL}~\cite{irl-ng00,irl-ziebart08}.
They introduce a discriminator $D_\phi$, parameterized by $\phi$, similar to the discriminator defined for GANs, but discriminating between actions taken by the expert and actions taken by the policy $\pi_\theta$ to be optimized. 
From a GAN perspective, the learned policy can be seen as the generator.
The GAIL objective is to find a saddle point of
\begin{align}
\mathcal{L}^{\mathrm{GAIL}}\left(\theta, \phi\right) &= \mathbb{E}_{\pi_\theta}\left[\log D_\phi\left(s_t,a_t\right)\right]+ \nonumber \\
&+\mathbb{E}_{\pi_E}\left[\log \left(1-D_\phi\left(s_t,a_t\right)\right)\right]-\lambda \mathcal{H}\left(\pi_\theta\right)
\label{eq:gail_objective}
\end{align} 
where $\mathcal{H}$ is the current policy's entropy, as already used for IRL, $\pi_\theta$ is the learned policy and $\pi_E$ is the expert policy. 
In practice, the expert policy is often represented by a dataset consisting of trajectories $\tau$. 
The authors suggest alternating gradient steps using Adam on $\phi$ to increase Eq.~\ref{eq:gail_objective} w.r.t. $D$ and TRPO~\cite{trpo-schulman15} on $\theta$ to decrease Eq.~\ref{eq:gail_objective} w.r.t. $\pi$. \cite{gail-ho16} 

As with humans, imitation is not only applicable to learn the behavior from an expert, but also to improve one's own behavior by reaffirming rewarding experiences made in the past and trying to achieve those again. 
However, besides the shared inspiration drawn from GANs, GAIL is dependent on (expert) demonstrations in order to learn a policy, while DIRECT is able to employ the concept of self-imitation on former beneficial trajectories in order to learn a policy, that maximizes the return of a given environmental reward function.

\paragraph{Self-Imitation Learning (SIL)} has been developed as an extension of A2C, adding a buffer $\mathcal{B}$ that stores trajectories $\tau$ previously encountered by the policy $\pi_\theta$. Thereby, the policy is additionally optimized towards beneficial recent behavior, i.e. trajectories with positive advantage values, or, where the actual return is greater than the estimated value. This extension has been shown to yield a deeper exploration of the reward landscape than comparable approaches by exploiting past experiences. \cite{sil-oh18}

Similarly to {experience replay}~\cite{acer-wang16}, {Oh et al.}~\shortcite{sil-oh18} propose to extend the A2C framework by a buffer $\mathcal{B}$ storing the policy's previous experiences to be revisited.
After each rollout, the buffer is extended by the recently encountered states, actions and cumulative rewards $\mathcal{B}=\left\{\left(s_t,a_t,G_t\right)\right\}$.
Policy and value estimate are updated sequentially towards both the RL objective by applying the on-policy A2C losses and past beneficial trajectories stored in the experience buffer by applying the following adapted A2C losses:
\begin{align}
\mathcal{L}^{\mathrm{SIL}} &= \mathbb{E}_{s_t,a_t,G_t\in \mathcal{B}}\left[ \mathcal{L}^{\mathrm{SIL}}_{policy} + \beta^{\mathrm{SIL}} \mathcal{L}^{\mathrm{SIL}}_{value} \right] \\
\mathcal{L}^{\mathrm{SIL}}_{policy} &= - \log \pi_\theta\left(a_t \mid s_t \right)\left( G_t - V_\theta\left(s_t\right) \right)_+ \\
\mathcal{L}^{\mathrm{SIL}}_{value} &= \frac{1}{2} \left\|  \left( G_t - V_\theta \left( s_t\right) \right)_+ \right\|^2
\end{align}
where $ \left(\cdot\right)_+=\max \left(\cdot, 0 \right) $, is constraining updates of both the policy and the value estimate to be only made for beneficial trajectories (i.e., where the actual return is greater than the estimated value).
Despite being formally designed as an extension of Actor-Critic methods, SIL has been shown to be also applicable to PPO. \cite{sil-oh18}

Overall, both DIRECT and SIL use beneficial previous experience to improve exploration. 
However, DIRECT constructs a single surrogate reward function that can be used for policy optimization.
In contrast, SIL incorporates those previous experiences in form of a replay buffer.
Thus, the policy is trained both with new and good previous samples, but there is no intermediate reward to incentivize the exploration of a sparse reward signal.

\section{Discriminative Reward Co-Training}

\begin{figure}[hbt]
\includegraphics[width=\linewidth]{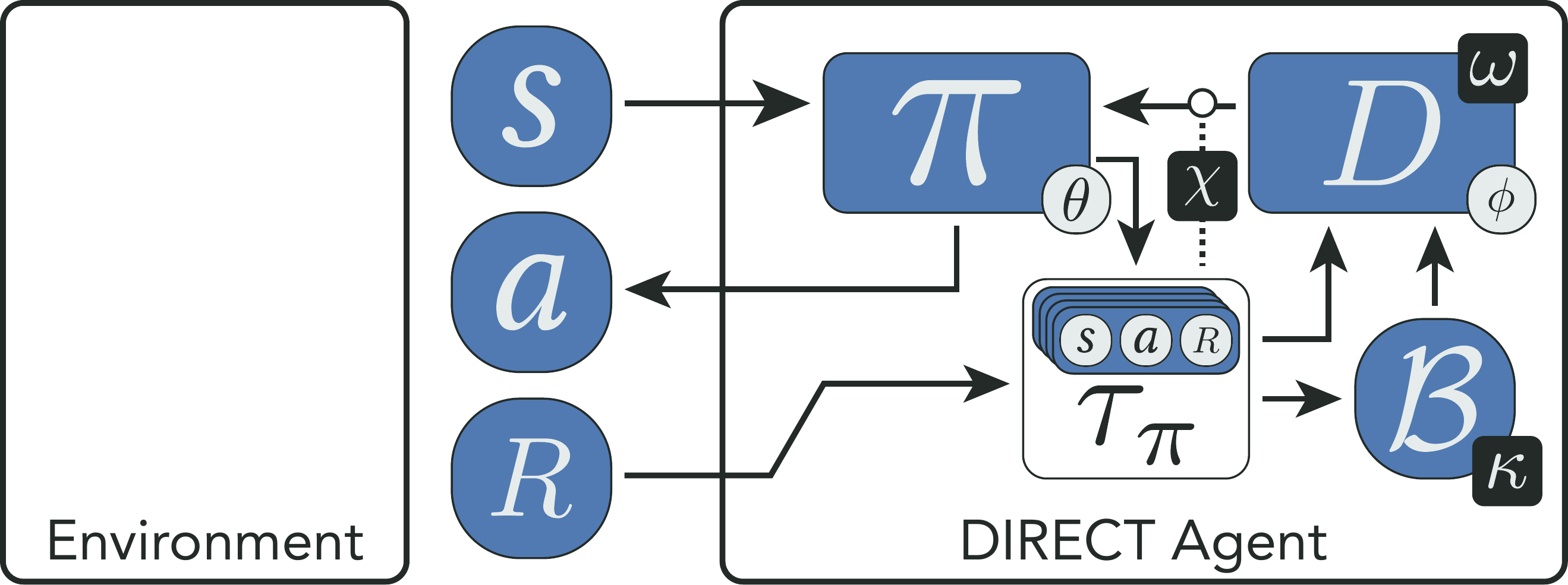} 
\caption{\textbf{DIRECT Architecture}
\textmd{Following policy $\pi$, the agent executes actions $a$ in the environment as a result of observing its state $s$, receiving a reward $r$ for the actions' consequences. 
Thereby, the policy generates trajectories $\tau_\pi$ containing states, actions and rewards. 
After every rollout, the return $G$ for each trajectory is calculated, and the self-imitation buffer $\mathcal{B}$ is updated. 
Storing the best $\kappa$ trajectories $\tau_\pi$, the discriminator $D$, parameterized by $\phi$, is then updated to differentiate between trajectories sampled from the buffer $\tau_\mathcal{B}$ and trajectories sampled from the current policy $\tau_\pi^*$. 
The update frequency of the discriminator in relation to the policy is given by $\omega$.
Ideally, the trajectories in $\mathcal{B}$ converge towards expert-like behavior during the training process. 
Therefore, the discriminator output can be used as a reward signal to update the policy parameters $\theta$, optimizing it towards good or even optimal past experiences. 
Additionally, the hyperparameter $\chi$ can be used to tune the mixture between the environmental reward and the discriminative reward.}
} 
\Description{Discriminative Reward Co-Training Architecture}
\label{fig:DIRECT}
\end{figure}

We propose an architecture called \textbf{Discriminative Reward Co-Training (DIRECT)}, as an extension of GAIL applied to self-imitation. 
\autoref{fig:DIRECT} illustrates the components of the architecture and their connection, which are formally introduced in the following section.
The overall DIRECT objective is defined as 
\begin{align}  \underset{\theta}{\mathrm{argmin}} \;\underset{\phi}{\mathrm{argmax}}\;\mathcal{L}^{\mathrm{DIRECT}}\left(\theta, \phi\right), with\end{align}
\begin{align} 
\mathcal{L}^{\mathrm{DIRECT}}\left(\theta, \phi\right) &= \mathbb{E}_{\tau_{E\sim{\mathcal{B}}}}\left[\log \left(1-D_\phi\left(s,a, G\right)\right)\right]+ \nonumber \\ 
    &+\mathbb{E}_{\tau_\pi}\left[\log D_\phi\left(s,a, G\right)\right]-\lambda \mathcal{H}\left(\pi_\theta\right)
\label{eq:DIRECT_objective}
\end{align}

\paragraph{Self-Imitation Buffer $\mathcal{B}$}
As suggested for self-imitation learning, we use a buffer storing beneficial past trajectories $\tau_\pi^*$ that the agent should reiterate. 
Beneficial state-action transitions are determined by their discounted return they achieved when executed in the environment. 
As illustrated in \autoref{fig:DIRECT}, the self-imitation buffer $\mathcal{B}$ is updated continuously while the policy $\pi_\theta$ is optimized and therefore generates new experience $\tau_\pi$. 
Thus, the trajectories in the buffer $\mathcal{B}$ are enabled to evolve, pushing the policy $\pi_\theta$ towards promising areas, and also might converge towards optimal behavior.
To survey this evolution of $\mathcal{B}$ and ensure its improvement towards optimal behavior, we introduce the \textit{buffer reward} as an additional reward, reflecting the mean reward of the trajectories $\tau_\mathcal{B}$ currently stored in the self-imitation buffer $\mathcal{B}$ . 
Also, we record the number of successful buffer updates, where old experience is replaced by improved and more recent behavior, we refer to as \textit{buffer momentum}.

\paragraph{Discriminator $D$}
Inspired by GAIL and GANs, the discriminator $D$ is trained to distinguish between samples generated by the current policy $\tau_\pi$ and buffer samples $\tau_E$. 
Applying these concepts to SIL, we do not require any expert behavior to be imitated. 
Thus, expert samples are replaced by samples drawn from the self-imitation buffer $\tau_\mathcal{B}$, as specified in \autoref{eq:DIRECT_objective}, the discrimination in DIRECT happens between former good experiences $\tau_\mathcal{B}$ and experiences by the current policy $\tau_\pi$.
As for GAIL, the discriminator's target is to label beneficial (expert) samples as ``1'' and samples generated by the current policy $\pi_\theta$ to be optimized as ``0''.
Therefore, beneficial behavior is rewarded by the discriminator $D$.  
As the samples in the self-imitation buffer $\mathcal{B}$ are supposed to converge towards optimal behavior throughout the training process, the discriminator $D$ should be able to guide the policy optimization towards an optimal solution.
In contrast to GAIL which imitates an expert, we attempt to maximize an environmental reward signal using self-imitation. 
We therefore propose adding the accumulated environmental return, already stored in the self-imitation buffer $\mathcal{B}$, as an input to the discriminator $D$.

From a GAN perspective, this addition is comparable to conditional GANs, adding labels to the training data \cite{cgan-mirza14}.
In contrast to GANs, however, we do not wish to have a fast converging discriminator $D$, as the self-imitation buffer $\mathcal{B}$, where target samples are drawn from, gradually evolves (hopefully towards optimal behavior) during the training process. 
Thus, a discriminator that converged prematurely, i.e., towards discriminating suboptimal samples as expert-like behavior, is not intended. 
Therefore, balancing the inner and outer optimization loops plays a more prominent role for training DIRECT than it does for GANs or GAIL. 

\paragraph{Policy Optimization}
Overall, DIRECT can be seen as an extension of the components described above, applicable to any model-free RL algorithm, providing an intermediate instance for processing the reward signal passed to the policy optimization instance (cf. \autoref{fig:DIRECT}). 
Similar to reward shaping approaches, we expect this extension to benefit the optimization by providing a smoother reward landscape that is easier to exploit by the agent. 
This guidance is especially helpful for sparse reward landscapes and hard exploration problems.
For optimizing the policy, we chose to use {Proximal Policy Optimization (PPO)}~\cite{ppo-schulman17}, an approach shown to be successfully applicable to a wide variety of RL problems. 
The reward signal used by PPO to calculate the return after every policy rollout as is calculated as follows: 
\begin{equation}
  r^*_t\left(s_t, a_t\right)=\chi D\left(s_t, a_t, G_t \right)+\left(1-\chi\right)r_{t+1},
  \label{eq:DIRECT_reward_out}
\end{equation} 
where $r_{t+1}$ is the real reward (\textit{environmental reward}) received by the agent from the environment for executing action $a_t$ and $\chi$ is a hyperparameter determining the reward mixture, such that for $\chi=0$, no discriminative reward is used and the algorithm  exactly resembles PPO. With $\chi=1$, PPO only receives a reward originating from the discriminator without any access to the real environmental reward signal. 

\paragraph{Training procedure}
To elucidate the proposed components, previously introduced and illustrated in \autoref{fig:DIRECT}, Algorithm~\ref{algo:DIRECT} shows the training procedure of the  proposed architecture for Discriminative Reward Co-Training. 
\begin{algorithm}[htb]
\caption{Discriminative Reward Co-Training}\label{algo:DIRECT}
\begin{algorithmic}
\State Initialize policy parameters $\theta$
\State Initialize discriminator parameters $\phi$
\State Initialize self-imitation buffer $\mathcal{B}\leftarrow\emptyset$
\For{each iteration}
    \State Sample policy trajectories $\tau_\pi$ by performing a PPO rollout
	\State $\tau_\pi \gets \left[\left(s_0,a_0,R_1 \right), \left(s_1,a_1,R_2 \right), \dots, \left(s_t,a_t,R_{t+1} \right)\right] $  
    \State Calculate partial returns $G(\tau)$ following Eq. \ref{eq:return}
	\State Update self-imitation buffer $\mathcal{B}$, prioritizing samples via $G(\tau)$:
    \State $\mathcal{B} \gets \left[\left(s_0,a_0,G(\tau) \right), \dots, \left(s_t,a_t,G(\tau) \right)\right] $ 
	\State Sample beneficial trajectories $\tau_\mathcal{B}$ from $\mathcal{B}$
	\State Update the discriminator parameters $\phi$, to maximize Eq. \ref{eq:DIRECT_objective}
	\State Calculate discriminative reward $r^*_t$ following Eq. \ref{eq:DIRECT_reward_out} 
	\State Update policy parameters $\theta$ to minimize Eq. \ref{eq:DIRECT_objective} 
\EndFor
\end{algorithmic}
\end{algorithm}

\section{Experimental Setup}
All implementations can be found here \footnote{\url{https://github.com/philippaltmann/DIRECT}}.

\begin{figure*}[!ht]
\centering
  \subfloat[Training]{ \label{fig:envs:DenseTrain}
   \includegraphics[width=0.28\textwidth]{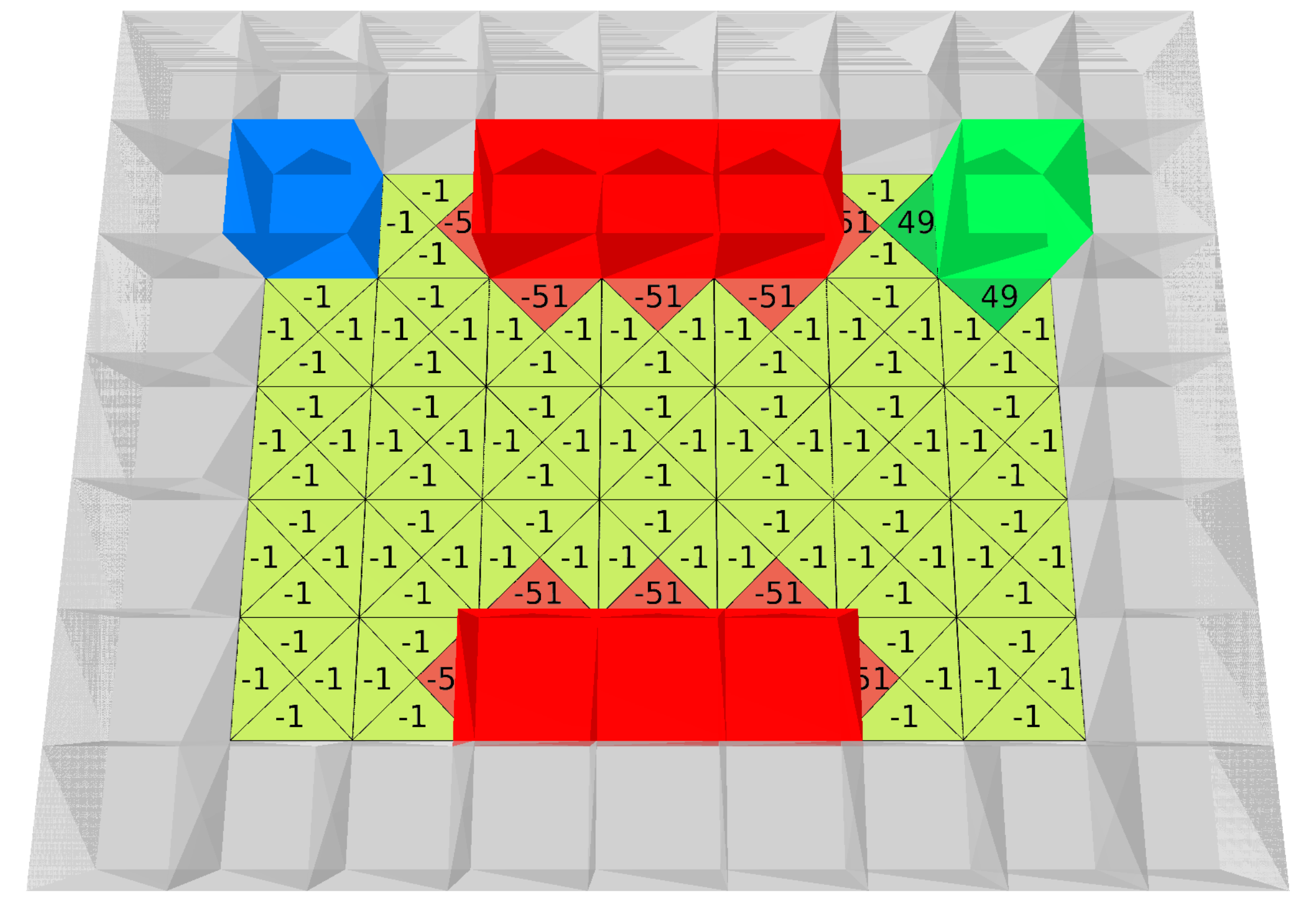}}
  \subfloat[Obstacle Shift]{ \label{fig:envs:DenseObstacleShift}
   \includegraphics[width=0.28\textwidth]{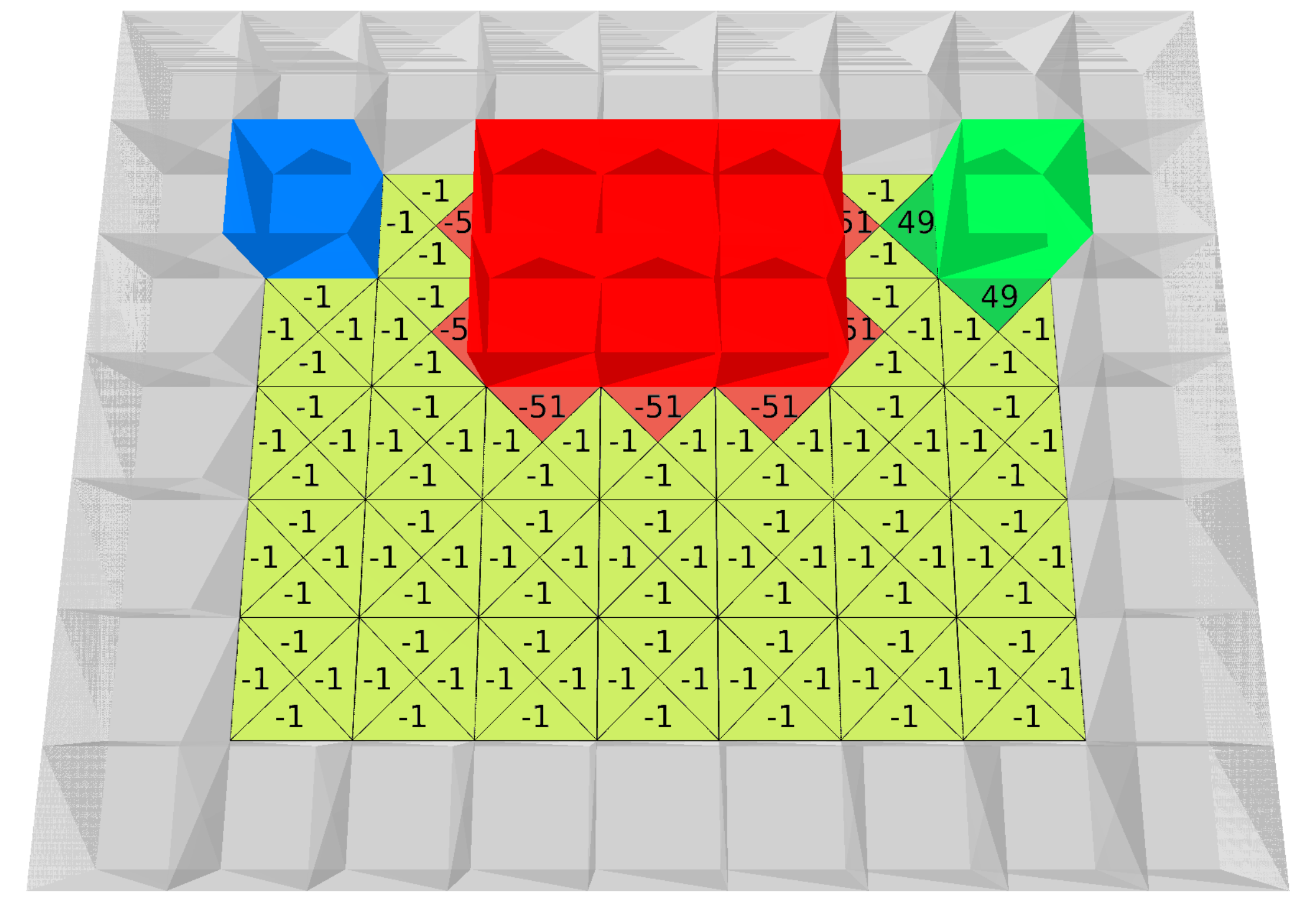}}
  \subfloat[Target Shift]{ \label{fig:envs:DenseTargetShift}
   \includegraphics[width=0.28\textwidth]{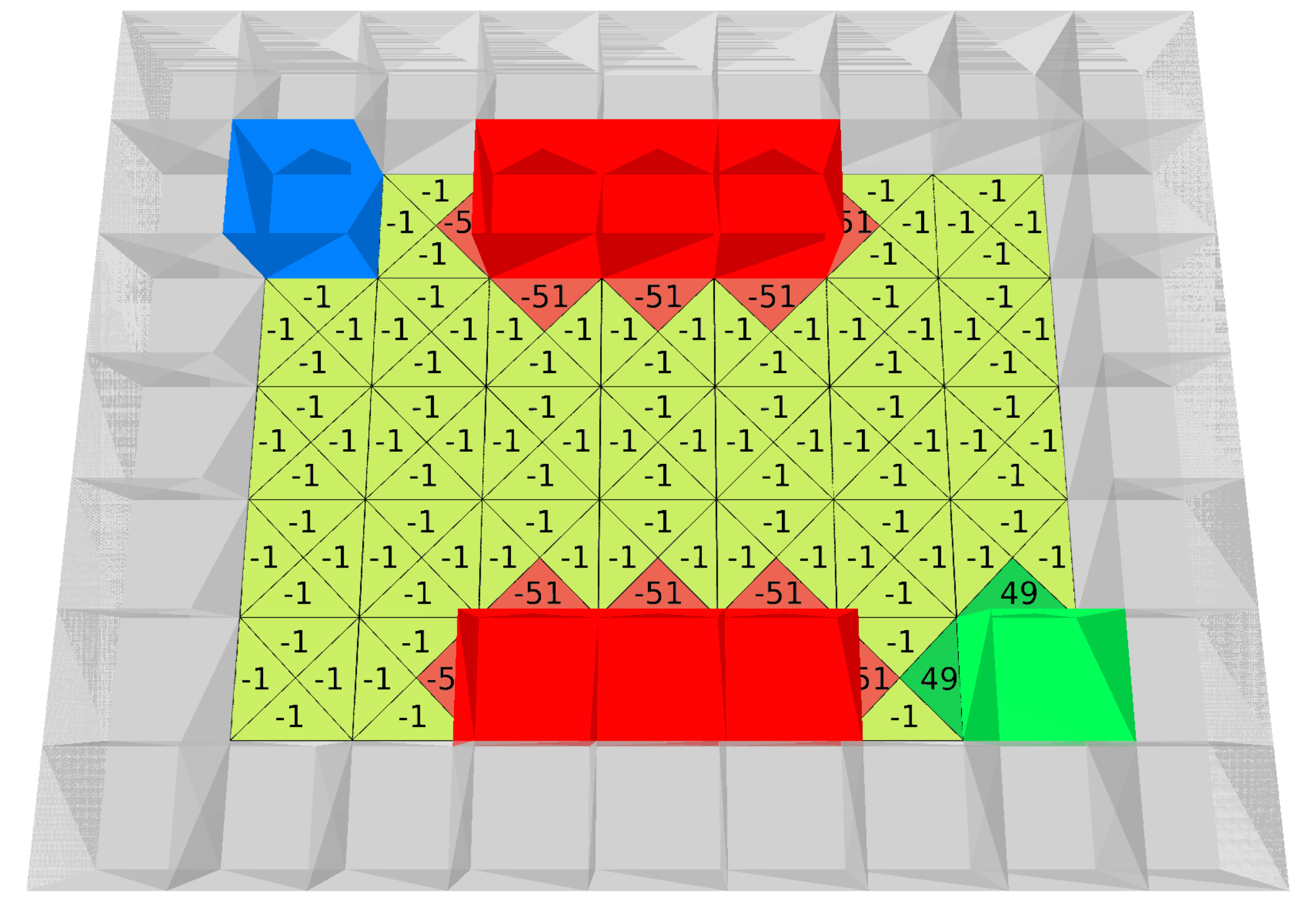}}
\caption[Dense Training]{Evaluation Environment Variations with dense reward landscape adapted from \cite{safegrid-leike17}. \textmd{The goal of the agent (blue) is to reach the target (green) rewarded without touching the lava (red). Both contact to lava and target fields terminate the episode with a reward of +50 and -50 respectively. To incentivize shorter paths each step is rewarded -1 which is  propagated to the agent directly in the dense reward setting and upon episode termination in the sparse reward setting.}}\label{fig:Environments}
\Description{Evaluation Environment Variations with dense reward landscape. The goal of the agent (blue) is to reach the target (green) rewarded without touching the lava (red). Both contact to lava and target fields terminate the episode with a reward of +50 and -50 respectively. To incentivize shorter paths each step is rewarded -1 which is  propagated to the agent directly in the dense reward setting and upon episode termination in the sparse reward setting.}
\end{figure*}

\paragraph{Hyperparameters}
Unless denoted otherwise we use DIRECT with the following hyperparameters that have shown the best results in preliminary experiments:
\begin{itemize}
    \item $\kappa=2048$: a self-imitation buffer $\mathcal{B}$ size, that resembles the rollout horizon of the used PPO implementation,
    \item $\omega=1/2$: in sparse reward settings resulting in concurrently training $D_\phi$ and the $\pi_\theta$, as the discriminator $D$ is updated with the samples from both the policy $\tau_\pi$ and the buffer $\tau_\mathcal{B}$,
    \item $\omega=2/1$: in dense reward settings, due to the well-defined reward signal a higher update rate $\omega$ allows for faster adaption of the discriminator $D$, thus faster overall training,
    \item $\chi=0.5$ for dense reward settings, to showcase the ability of DIRECT to create an homogeneous reward signal, and 
    \item $\chi=1.0$ for sparse reward setting, to assess the prospects of a purely discriminative reward signal. 
\end{itemize}

To benchmark DIRECT we use the PPO, A2C and DQN implementations by \cite{stable-baselines3} with default PPO and DQN hyperparameters and A2C hyperparameters as suggested by \cite{safegrid-leike17}, as well as a SIL implementation according to \cite{sil-oh18}.

\paragraph{Domain}
We chose to evaluate the DIRECT using environments from the AI Safety Gridworlds \cite{safegrid-leike17} adapted to OpenAI gym \cite{openai-gym}. 
The environments used are shown in \autoref{fig:Environments}.
Training a policy in the training environment (cf. \autoref{fig:envs:DenseTrain}) and evaluating the policy within the Obstacle- and Target Shift environments (cf. \autoref{fig:envs:DenseObstacleShift} and \autoref{fig:envs:DenseTargetShift}) reveals the approaches' robustness against distributional shift, which is considered one central problem to be solved for safe RL \cite{safegrid-leike17, safety-amodei16}.
Beside this concisely crafted challenges, we chose to evaluate DIRECT using the safety gridworlds, as they allow for better interpreability and visualizations of resulting policies and the course of training. 

To assess the ability of DIRECT to provide a surrogate direction to the policy optimization, we furthermore added a sparse reward setting, where the agent receives zero rewards throughout the episode and a final non-zero reward at terminal states.

Besides testing an agents exploratvie capabilites and robustness against changes in the environment, the given environments are also suitable for evaluating an agent's \textit{safe exploration} capabilities, with the lava fields representing a safety risk to be avoided, by analyzing the cause of termination. 
Agents that cause more failure terminations by running into lava fields, either during training or within unseen environments are arguably less safe compared to agents terminating their episodes as a result of reaching the maximum number of steps \cite{safety-amodei16}.
Terminating in the target state using the optimal path yields respective returns of 42, 40 and 40. 
In any case, reaching the target state yields a return of at least -52. 

\paragraph{Metrics}

As a performance metric for a policy's behavior, we use the average environmental return $G(\tau)$ over 100 episodes (mean return). 
In order to reduce the necessary compute, this metric is also used to terminate the training upon reaching a threshold of 95\% of the optimal solution, i.e. a return of 40 in environment (\ref{fig:envs:DenseTrain}), and a return of 38 in environments (\ref{fig:envs:DenseObstacleShift}) and (\ref{fig:envs:DenseTargetShift}).

To assess the resulting policies, we run single episode, deterministic evaluations and record the mean return, as well as the episode termination reason of the trained policies.
All of the following results are averaged over eight runs with randomly selected seeds, trained in parallel in four environments for a maximum of 1 million timesteps.

\section{Training}\label{sec:training}

\begin{figure}[b]
  \centering 
  \includegraphics[width=\linewidth]{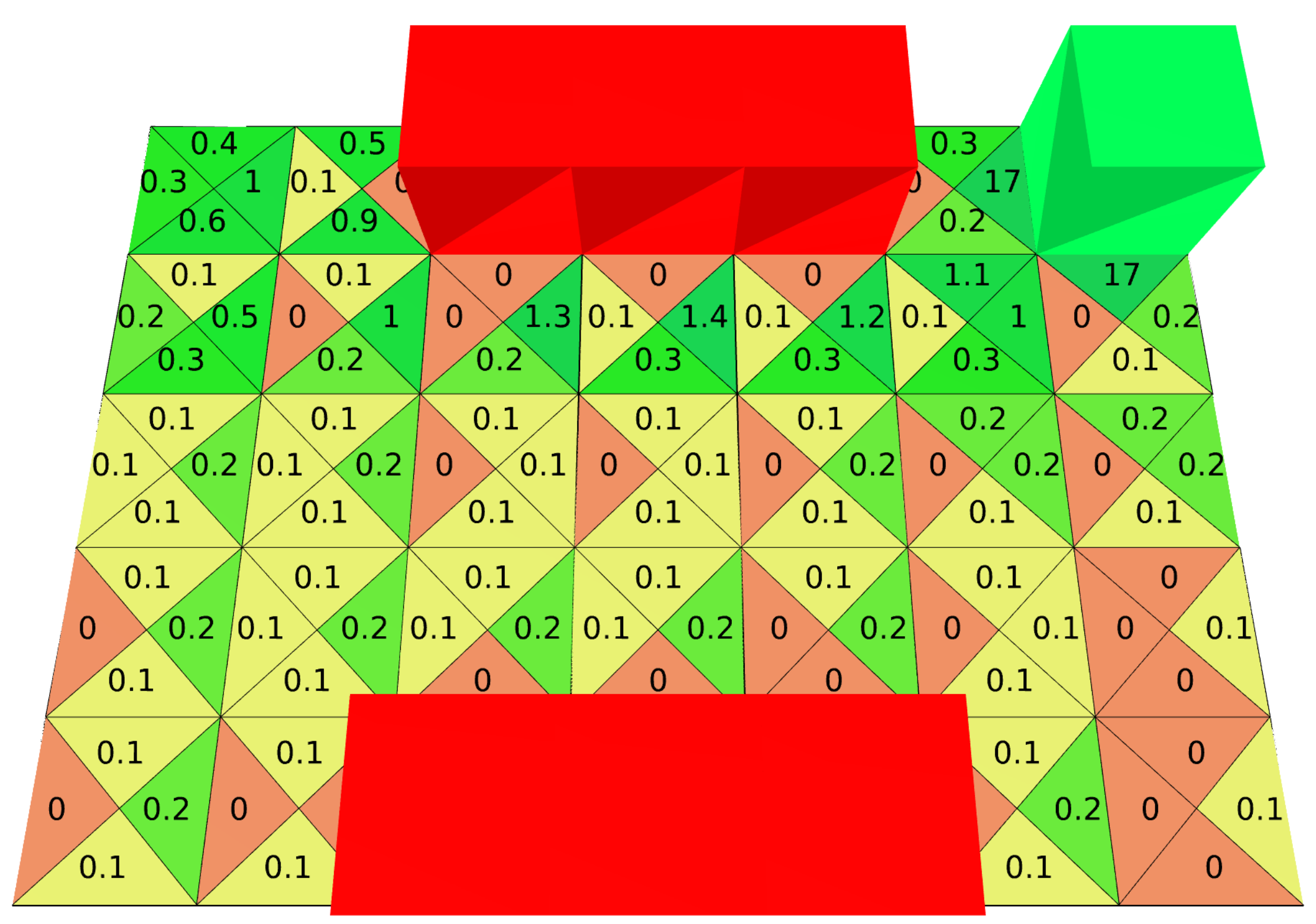}
  \caption{ DIRECT Reward Heatmap: \textmd{Directive reward on average over all runs after training in the Dense Reward Environment}}
  \label{fig:heatmap:direct-train-dense}
\Description{DIRECT Reward Heatmap after training in the Dense Reward Environment}
\end{figure}

\begin{figure}[t]
  \centering 
   \subfloat[Dense Reward]{\includegraphics[width=\linewidth]{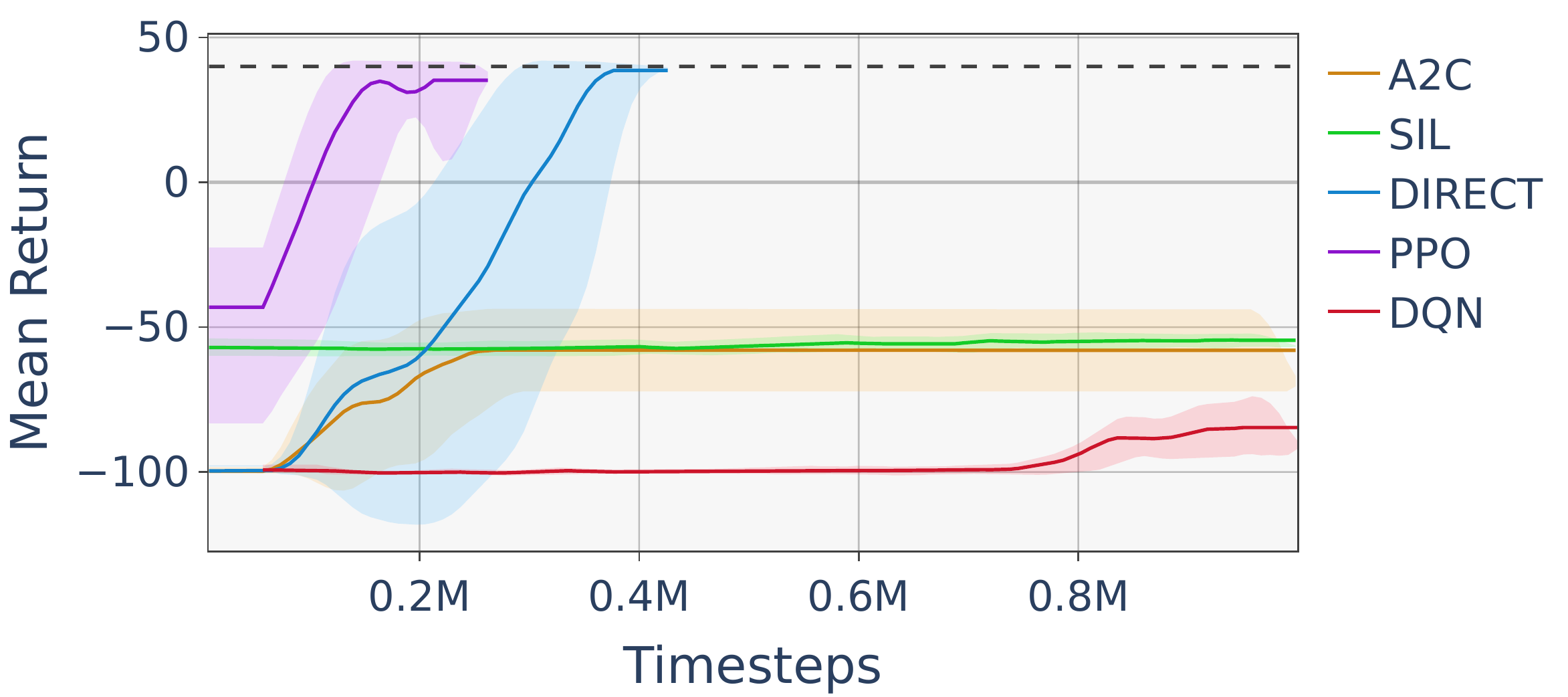}\label{fig:bench:training:dense}}\\
   \subfloat[Sparse Reward]{\includegraphics[width=\linewidth]{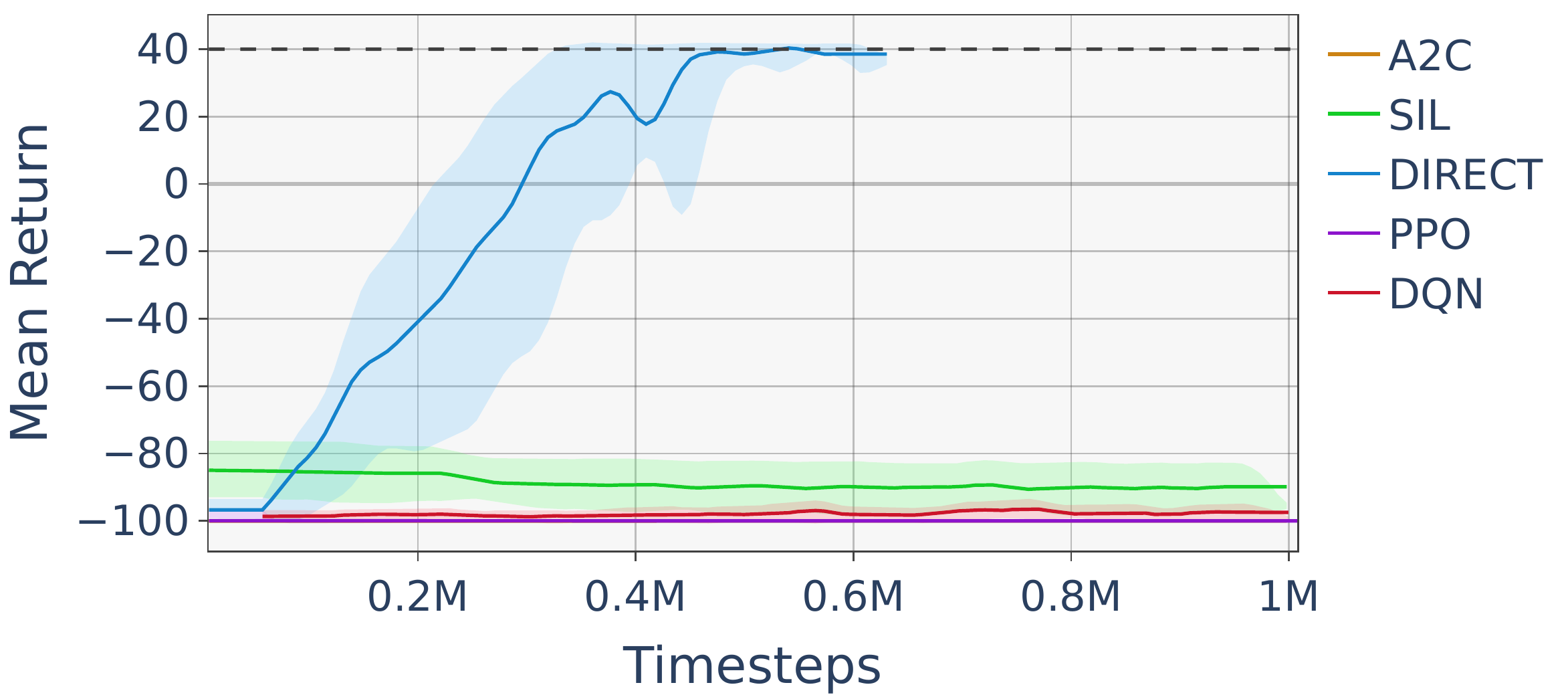}\label{fig:bench:training:sparse}}\\
  \caption{ Training Progress within the dense \hyperref[fig:bench:training:dense]{\textbf{(a)}} and sparse \hyperref[fig:bench:training:sparse]{\textbf{(b)}} reward environments: \textmd{ the optimization progress for A2C (orange), SIL (green), DIRECT (blue), PPO (purple), and DQN (red) averaged over eight runs, with the number of steps taken in the environment on the x-axis and the mean return on the y-axis. The shaded areas mark the 95\% confidence intervals, the reward threshold of 40 is displayed by the dashed line.}}
  \label{fig:bench:training}
\Description{Training progress in dense and sparse training environments}
\vspace{-1em}
\end{figure}

According to \autoref{fig:bench:training:dense}, both PPO and DIRECT are able to exploit the dense reward signal and find optimal policies within 200k and 350k timesteps on average respectively. 
The delay of convergence between PPO and DIRECT is most likely attributed to the double optimization loop architecture, where first, the policy needs to generate valuable experience to fill the buffer, before the discriminator can be conditioned to produce a valuable reward signal. 
\autoref{fig:heatmap:direct-train-dense} shows the average discriminative returns after training for all possible state-action combinations as in form of a heatmap.
Especially when comparing this to the environmental reward structure in \autoref{fig:envs:DenseTrain} the directive power of DIRECT can be seen. 
In addition to providing guidance toward the target, the directive instance has also learned to avoid lava fields, indicated by the zero rewards for all adjacent actions. 
However, even though the reward signal is well-defined and dense, both A2C and SIL get stuck in a local optimum around a return of -60.
Policies within this reward range are not able to reach the target, but cause an early episode termination by running into a near lava field. 
Thus, both SIL and A2C are not able to produce any optimal policies, which might be attributed to their short rollout horizon. 
DQN shows even worse performance, only reaching average return areas of -90 in the late optimization. 
The progress suggests that DQN might need a higher number of training steps in order to learn an optimal policy.

Overall, DIRECT shows competitive performance in comparison to the selected baselines.   
However, the real advantage of DIRECT really becomes visible looking at the training process of the sparse environment in \autoref{fig:bench:training:sparse}. 
While all other algorithms do not even converge to the locally optimal behavior of terminating the episode by running into the closest lava field, but produce policies that yield average returns of about -100, \textbf{all} Discriminative Reward Co-Trained policies converge to a near optimal behavior within under 650k timesteps of training. 
We attribute this superiority to the directive abilities of the discriminiative reward signal fostering improved exploration capabilities of valuable experience. 
By employing the concept of self-imitation DIRECT improves the optimization's sample efficiency, i.e. requiring to experience fewer good episodes in order to exploit optimal behavior. 

\begin{figure}[hbt]
  \centering 
    \subfloat[Episode Returns]{\includegraphics[width=\linewidth]{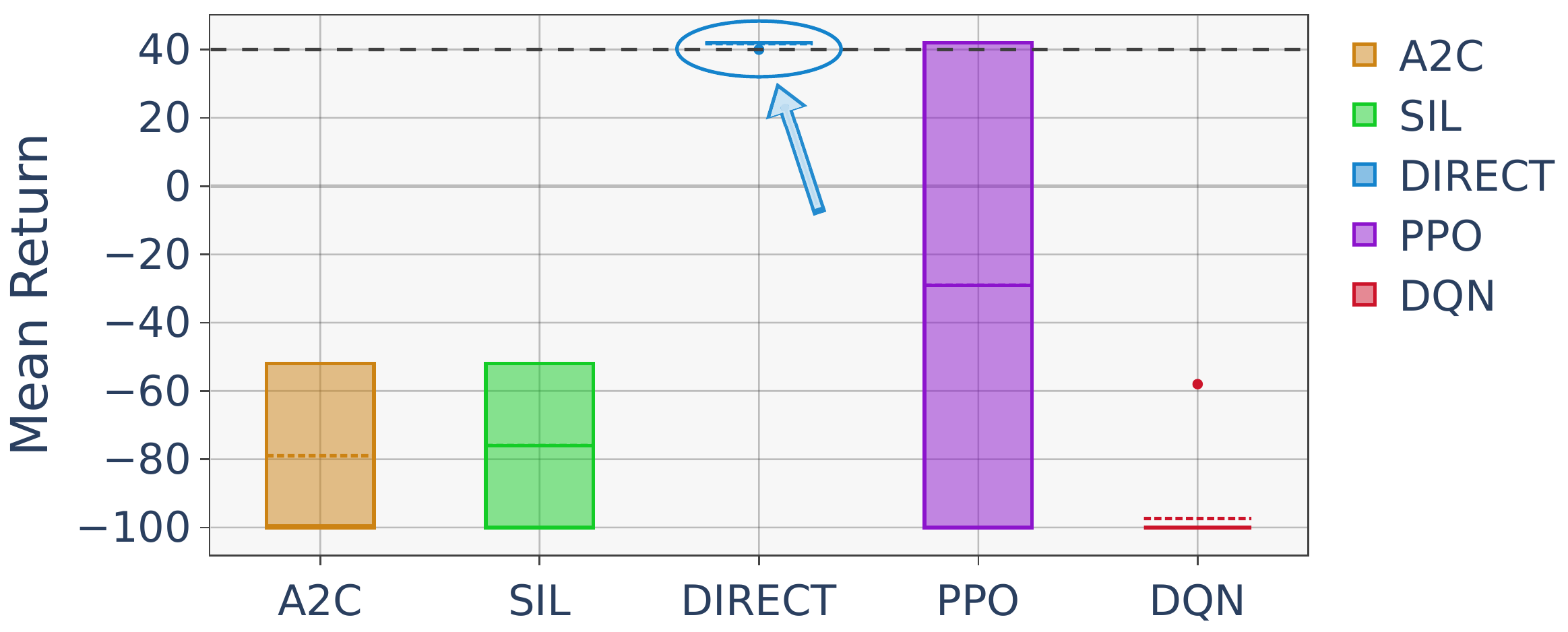}\label{fig:validation:training:return}}\\
    \subfloat[Episode Termination: \textmd{Percentage of episodes in which the agent reaches the target (green), hits an obstacle (red) or runs into a timeout (yellow).}]{\includegraphics[width=\linewidth]{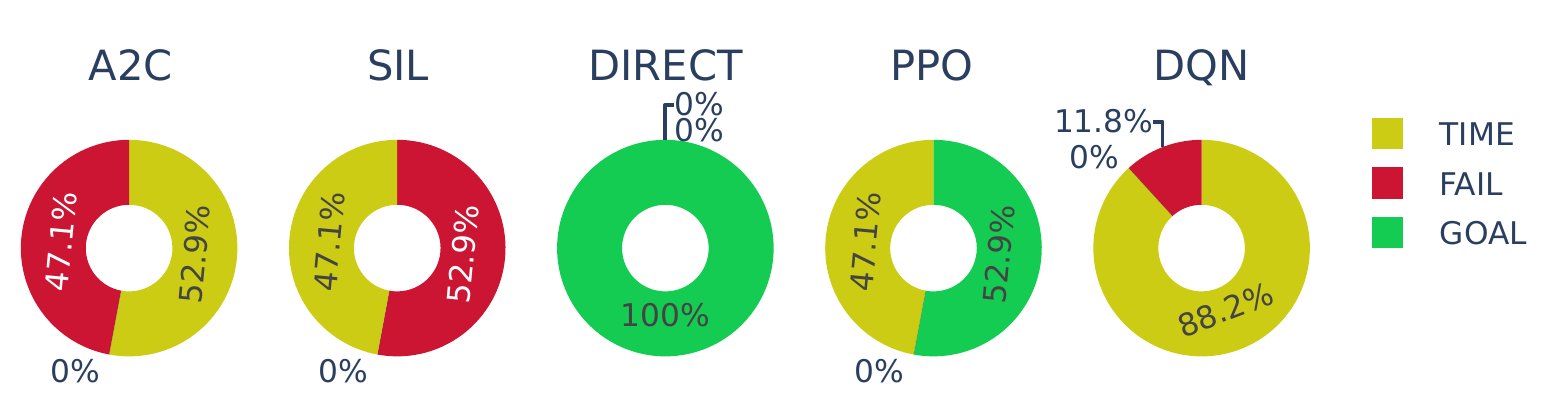}\label{fig:validation:training:termination}}\\
    \caption{ Training Evaluation: \textmd{Episode Returns (\hyperref[fig:validation:training:return]{a}) and termination reasons (\hyperref[fig:validation:training:termination]{b}) of policies trained with A2C, SIL, DIRECT, PPO, and DQN during deterministic evaluation in both the sparse and dense reward settings of the training environments (cf. \autoref{fig:envs:DenseTrain})}}
  \label{fig:validation:training}
\Description{Deterministic Training Evaluation: Deterministic mean returns and episode termination reasons}
\end{figure}

The overall training performance of all evaluated approached in both sparse and dense reward settings is summarized in \autoref{fig:validation:training}.
With all returns above the reward threshold, i.e. reaching the target via the optimal path or at most two more steps, Discriminative Reward Co-Trained polices outperform all all compared approaches. 
PPO-trained polices perform second-best, with half of the polices reaching into reward regions, where the target is reached. 
The worse performance of the remaining policies is attributed to PPO's bad ability to learn from sparse reward signals.  
A2C-, SIL- and DQN-trained policies did not converge to a target-finding behavior at all.
This superiority of DIRECT policies is also reflected in their safety where all (100\%) of the evaluation episodes terminate in the target state and, in contrast to all compared benchmarks, no unsafe behavior, i.e. failure terminations, is encountered. 

\section{Adaptation}
In the following section we will evaluate the approaches' robustness to distributional shift.
Therefore, we analyze their performance environments where, in comparison to the training environment (cf. \autoref{fig:envs:DenseTrain}), either the obstacles (cf. \autoref{fig:envs:DenseObstacleShift}), or the position of the target (cf. \autoref{fig:envs:DenseTargetShift}) has shifted. 
Again, with real-world applications in mind, we will consider both dense- and sparse reward settings. 
Due to its directing modifications to the reward signal, we expect DIRECT to yield policies with improve stability and adaptability even to shifting reward signals. 

\paragraph{Shifted Obstacles}

\begin{figure}[b]
  \centering 
  \includegraphics[width=\linewidth]{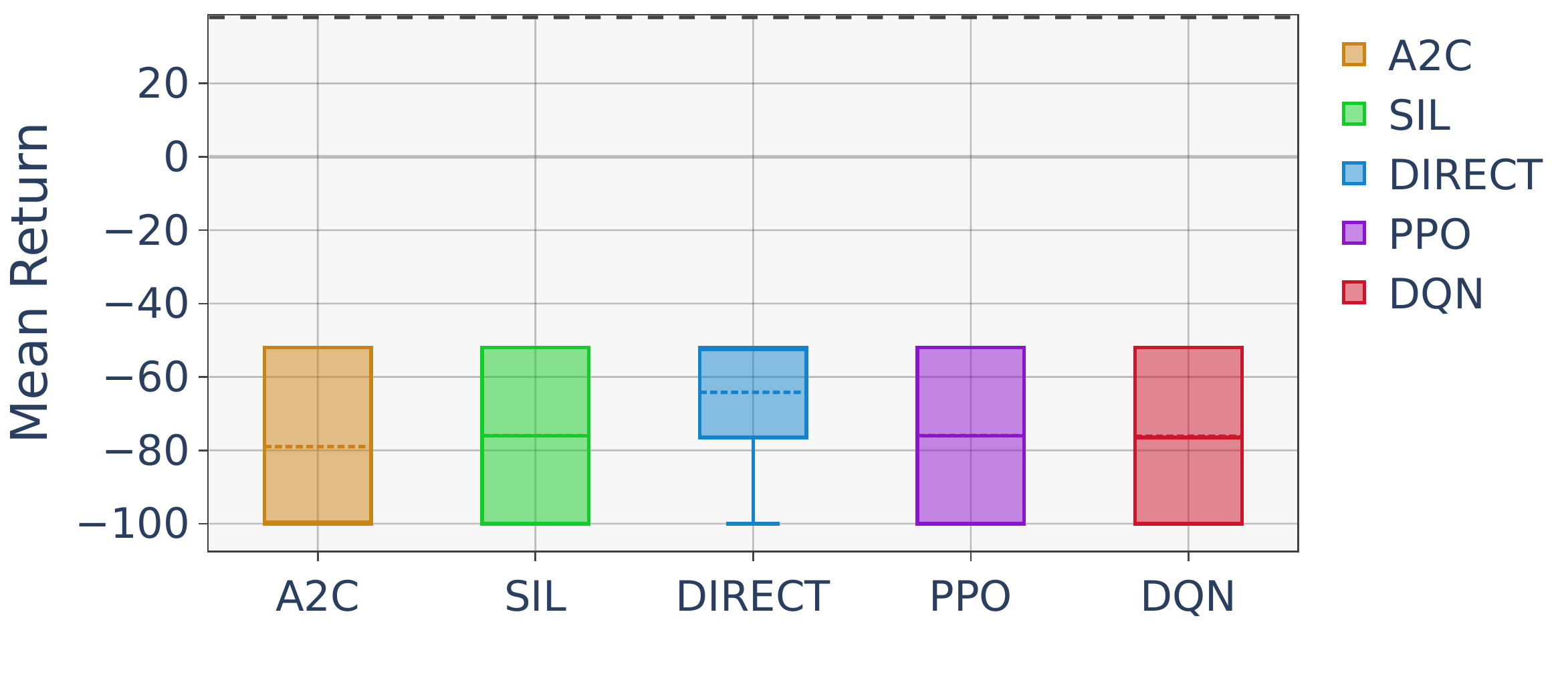}
  \caption{ Shifted Obstacle Evaluation: \textmd{Episode Returns of policies trained with A2C, SIL, DIRECT, PPO, and DQN during deterministic evaluation in both the sparse and dense reward settings of the shifted obstacle environments (cf. \autoref{fig:envs:DenseObstacleShift})}}
  \label{fig:validation:trained-obs}
\Description{Performance validation of trained policies in observation shift environment}
\end{figure}
In this evaluation scenario, the target position is the same as during training. However, the path that was shortest during training now is blocked by lava.
Thus, the agent needs to adapt its behavior to reach the target successfully. 

To asses the policies robustness to a distributional shift of its surrounding environment, we evaluated all previously trained policies deterministically in the shifted obstacle environment (cf. \autoref{fig:envs:DenseObstacleShift}).
The according results are shown in \autoref{fig:validation:trained-obs}.
All approaches achieve mean returns between -100 and -50.
This again highlights the susceptibility of RL algorithms to distributional shifts.
Overall, no trained policy is able to find a way around the re-positioned obstacles ad-hoc.
Therefore, we decided to perform a refinement-training on the previously trained polices to assess the underlying algorithms' adaptability to changes within the environment.

\begin{figure}[t]
  \centering 
   \subfloat[Dense Reward]{\includegraphics[width=\linewidth]{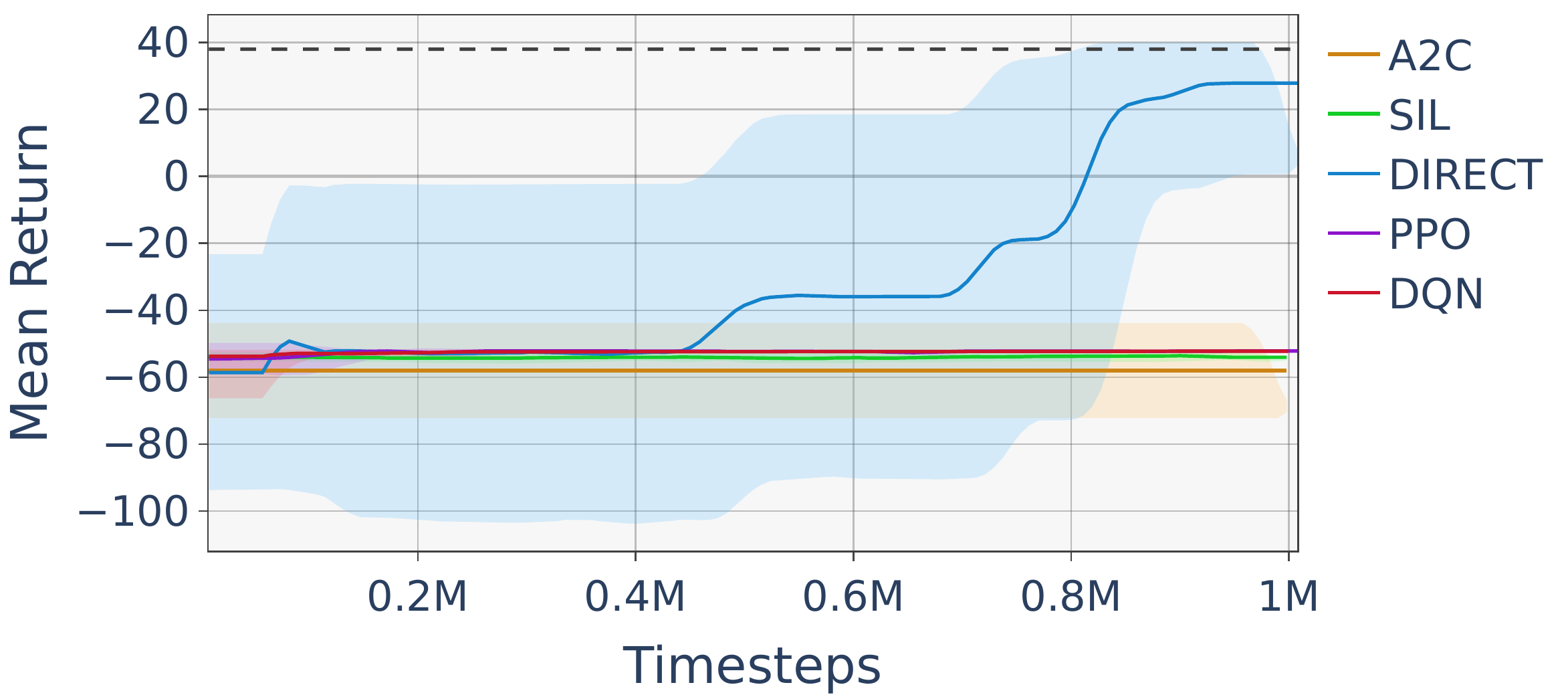}\label{fig:bench:adapt:obs:dense}}\\
   \subfloat[Sparse Reward]{\includegraphics[width=\linewidth]{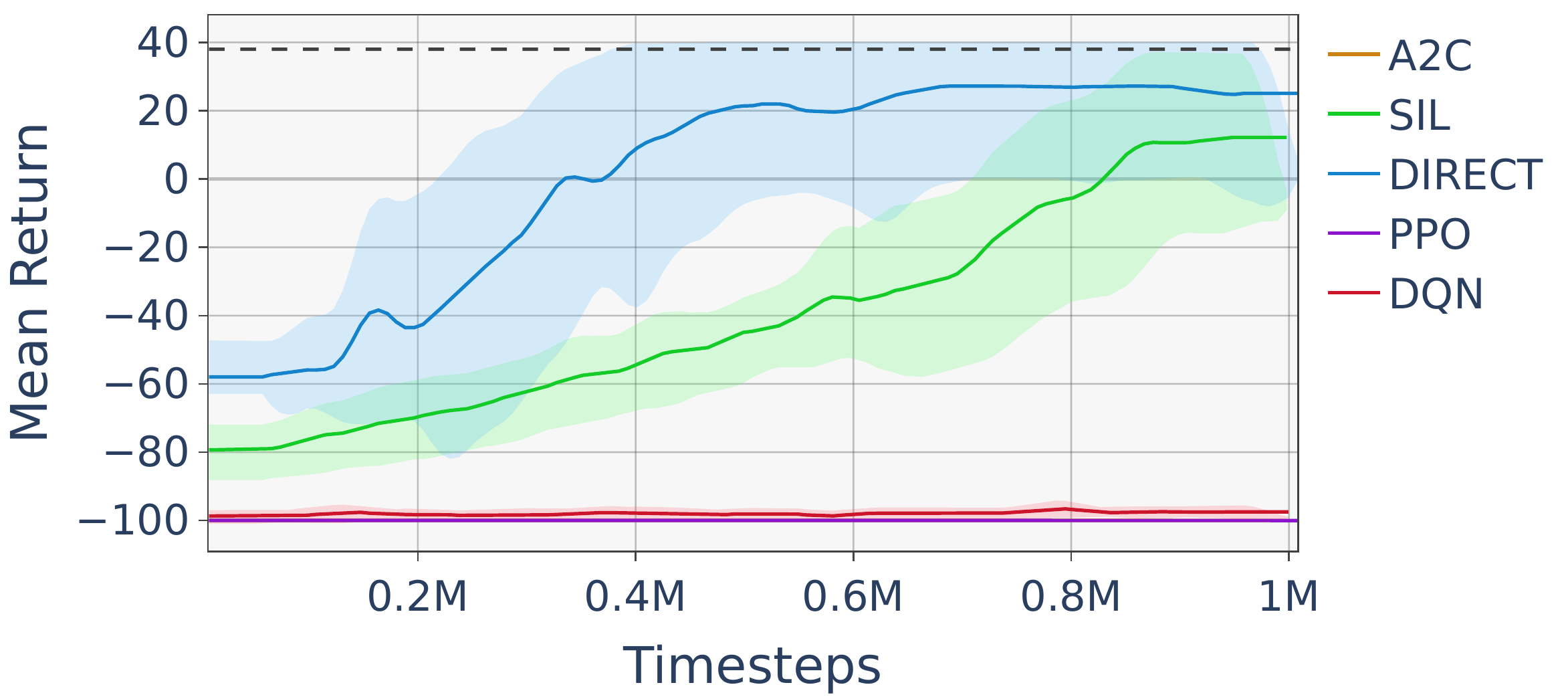}\label{fig:bench:adapt:obs:sparse}}\\
  \caption{ Adaption Progress Shifted Obstacle: \textmd{ the adaptation progress to dense \hyperref[fig:bench:adapt:obs:dense]{\textbf{(a)}} and sparse \hyperref[fig:bench:adapt:obs:sparse]{\textbf{(b)}} reward settings averaged over eight runs, with the number of additional timesteps taken in the shifted environment on the x-axis and the mean return on the y-axis. The shaded areas mark the 95\% confidence intervals, the reward threshold of 38 is displayed by the dashed line.}}
  \label{fig:bench:adapt:obs}
\Description{Training progress adapting previously trained policies to the obstacle shift environment}
\end{figure}

Trying to adapt to the dense reward signals (cf. \autoref{fig:bench:adapt:obs:dense}), all algorithms converge to sub-optimal returns of around -50 within the first 400k timesteps. 
However, in contrast to all other approaches, DIRECT is able to explore valuable regions of the reward landscape and exploit them via self-imitation within the following 600k timesteps, yielding a mean return of around ten. 
The mean return indicates that DIRECT is not able to find an optimal policies during the refinement training. 
However, in contrast to all other approaches, DIRECT is able to solve the scenario by reliably finding the target within 100 timesteps and without touching the lava.

\begin{figure}
  \centering 
    \subfloat[Episode Returns]{\includegraphics[width=\linewidth]{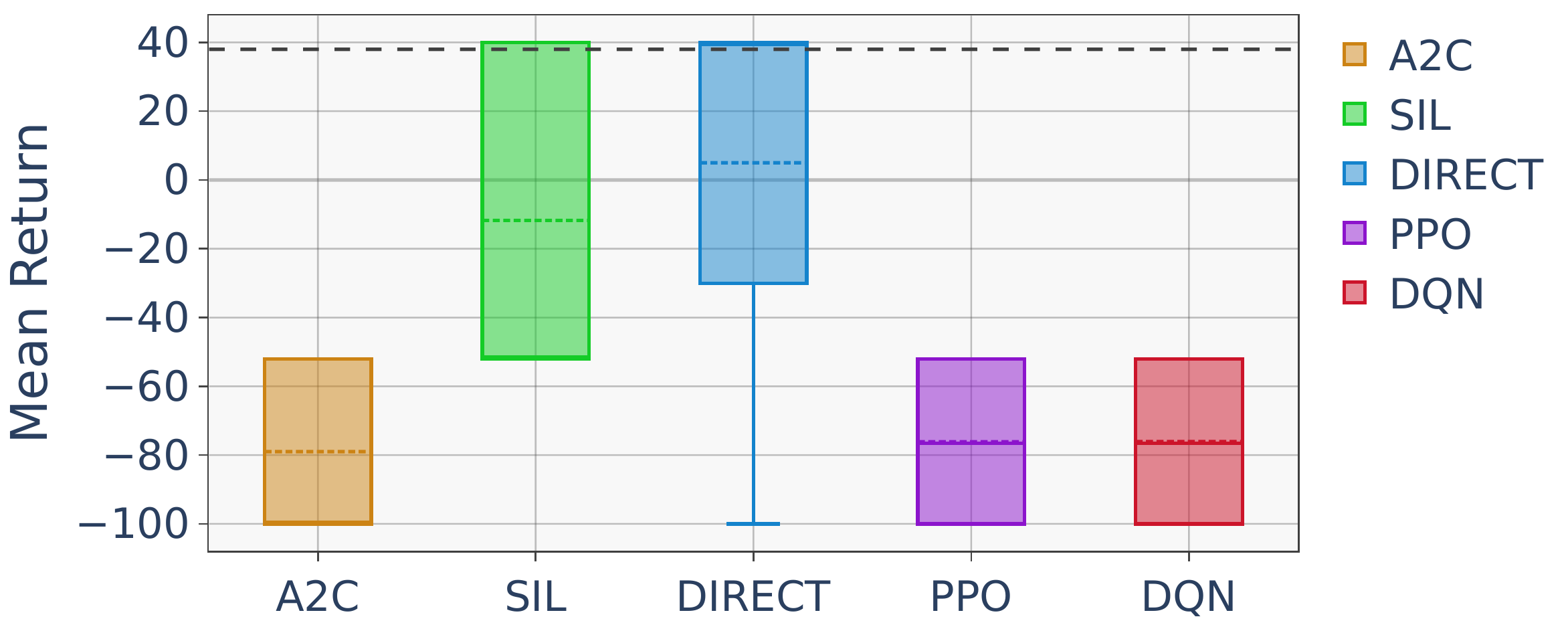}\label{fig:validation:adapt:obs:return}}\\
    \subfloat[Episode Termination: 
    ]{\includegraphics[width=\linewidth]{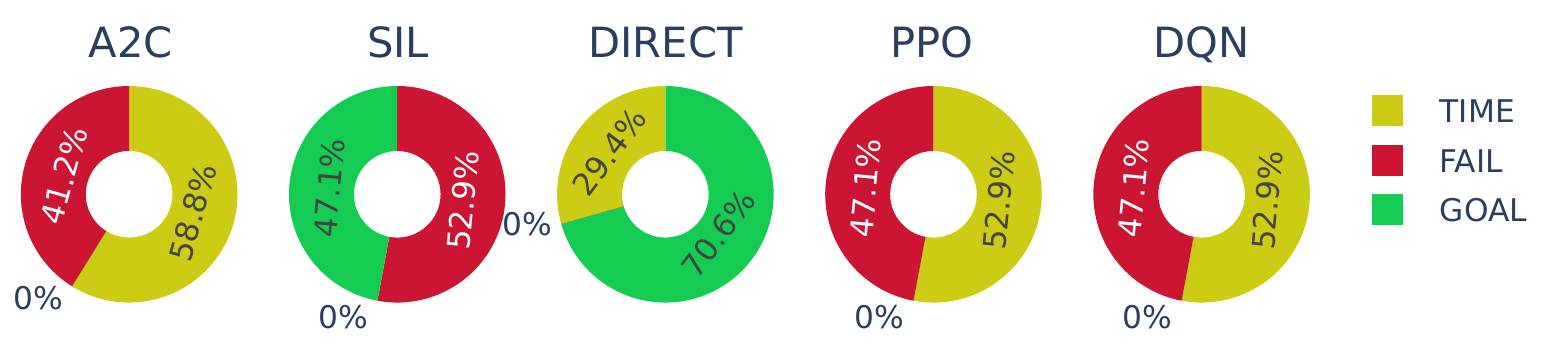}\label{fig:validation:adapt:obs:termination}}\\
    \caption{ Adaption Evaluation: \textmd{Episode Returns (\hyperref[fig:validation:adapt:obs:return]{a}) and termination reasons (\hyperref[fig:validation:training:termination]{b}) of policies trained with A2C, SIL, DIRECT, PPO, and DQN during deterministic evaluation in both the sparse and dense reward settings of the obstacle shift environments (cf. \autoref{fig:envs:DenseObstacleShift})}}
  \label{fig:validation:adapt:obs}
\Description{Performance validation of adapted policies in observation shift environment}
\end{figure}

Switching to the sparse reward setting (cf. \autoref{fig:bench:adapt:obs:sparse}), the performance of most approaches drops to -100, where the episodes most likely end with a timeout after 100 steps. 
Besides DIRECT, SIL shows surprisingly good adaptability to the shifted obstacle, converging to an average return of around ten throughout the training process.
This is especially surprising as the originating policies were never able to adequately solve the training environment.
However, DIRECT adapted policies again show even faster convergence towards high reward regions. 
Interestingly, despite the increased difficulty due to the sparse reward signal, DIRECT shows even better performance than within the less challenging dense reward setting. 
Overall, evaluating the adapted policies deterministically (cf. \autoref{fig:validation:adapt:obs}), only approaches incorporating self-imitation are able to adapt to the environmental shift. 
In concurence to previous work, this finding demonstrates the superiority of self-imitation based approaches in sample efficient exploitation of valuable experience \cite{sil-oh18}.
Again, in comparison, DIRECT policies yield superior safety avoiding lava fields overall.
All remaining approaches seem to suffer from base-policies that are already set to the prior training environment, and therefore are less adaptable to the changes within the environment.

\paragraph{Shifted target}
Complementary to the environment inspected above, the shifted target environment  (c.f. \autoref{fig:envs:DenseTargetShift}) contains a different target position compared to the original training environment but the obstacle positions remain unaltered. 
Thus, the agent also needs to adapt its behavior.
Again, we evaluated the policies trained in \autoref{sec:training} within the shifted environment. 
As the results resemble the findings of \autoref{fig:validation:trained-obs}, this plot is omitted.

\begin{figure}[htb]
  \centering 
   \subfloat[Dense Reward]{\includegraphics[width=\linewidth]{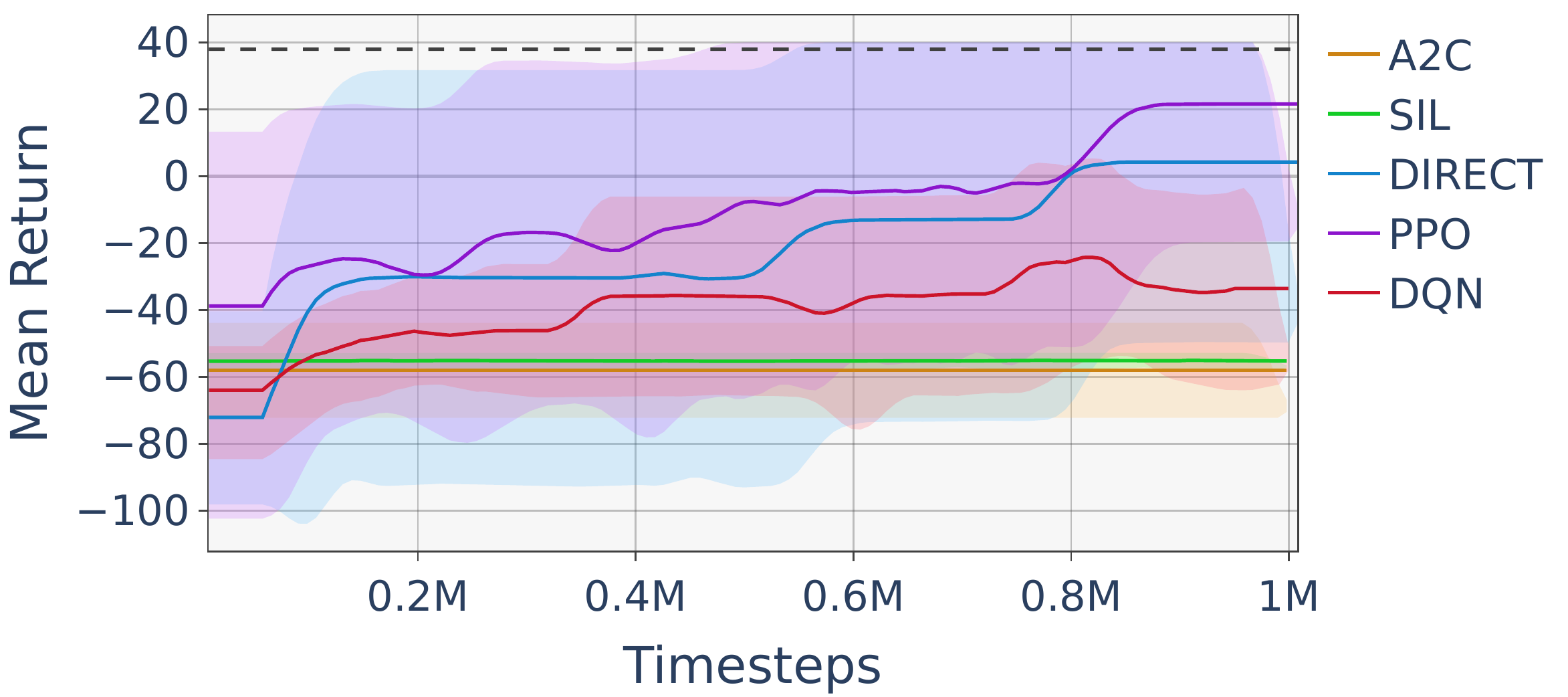}\label{fig:bench:adapt:target:dense}}\\
   \subfloat[Sparse Reward]{\includegraphics[width=\linewidth]{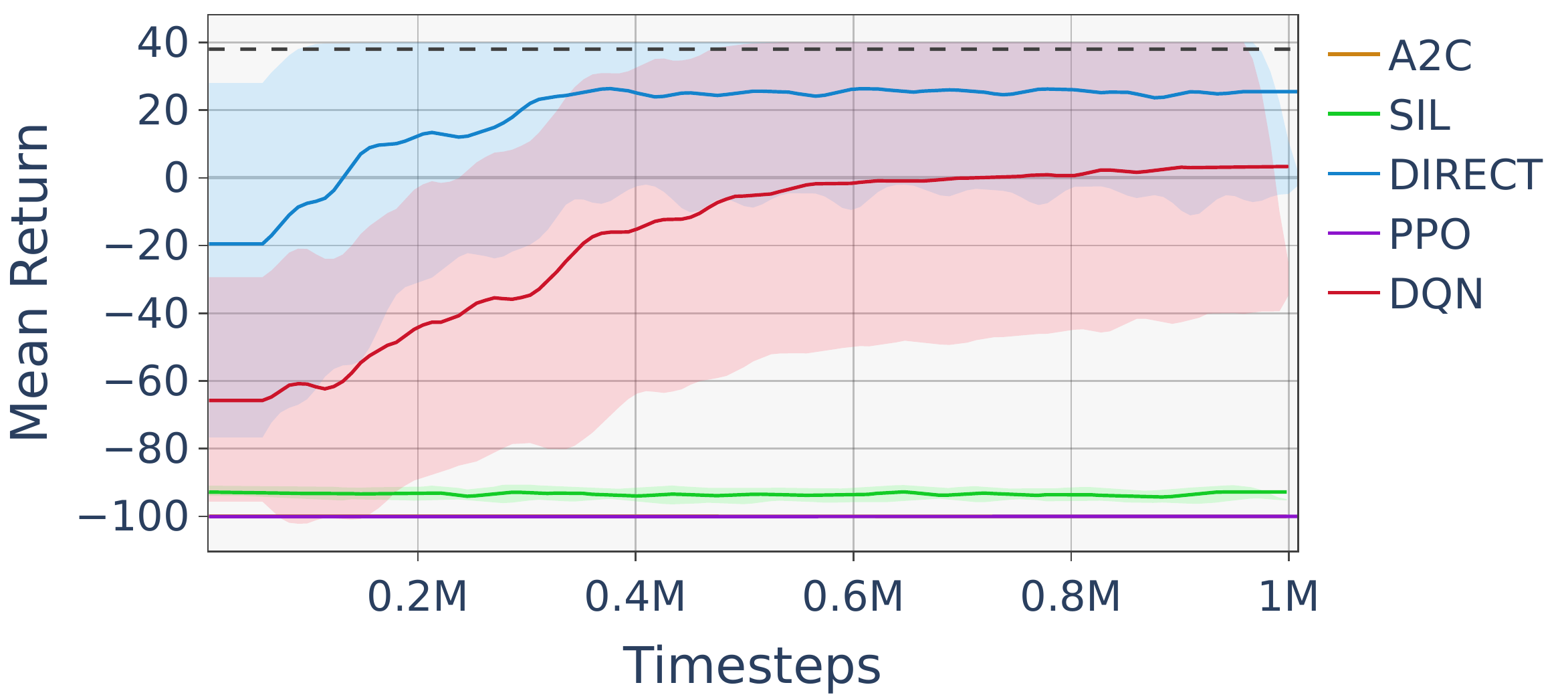}\label{fig:bench:adapt:target:sparse}}\\
  \caption{ Adaption Progress to the dense \hyperref[fig:bench:adapt:target:dense]{\textbf{(a)}} and sparse \hyperref[fig:bench:adapt:target:sparse]{\textbf{(b)}} shifted target (cf. \autoref{fig:envs:DenseTargetShift}) environments: \textmd{ the optimization progress averaged over eight runs, with the number of timesteps taken in the environment on the x-axis and the mean return on the y-axis. The shaded areas mark the 95\% confidence intervals, the reward threshold of 40 is displayed by the dashed line.}} 
  \label{fig:validation:adapt:target}
\Description{Training progress adapting previously trained policies to the target shift environment}
  
\end{figure}

Looking at the adaptation progress within the dense reward setting in \autoref{fig:validation:adapt:target}, similar to the training, both PPO and DIRECT reach areas of the return, where the target is reached. 
Again, presumably due to the delayed learning caused by the discriminator optimization, DIRECT converges slightly slower. 
Furthermore, DQN shows convergence toward target-reaching return regions, proving above hypothesis that its insufficient convergence within the training environment can be overcome by further training.

This is even strengthened within the sparse reward setting shown in \autoref{fig:bench:adapt:target:sparse}, where DQN shows second-best performance. 
Again, the performance of stuck approaches drops to a mean return of -100, while DIRECT is able to provide guidance towards the shifted target and yields the fastest convergence and best performing policies.

Overall DIRECT trained polices showed the best adaptability to both shifted targets and shifted obstacles. 
They exceed state-of-the-art approaches especially in sparse reward setting. 
However, in contrast to the initial training, none of the approaches yields an optimal policy.

\begin{figure}[htb]
  \centering 
  \includegraphics[width=\linewidth]{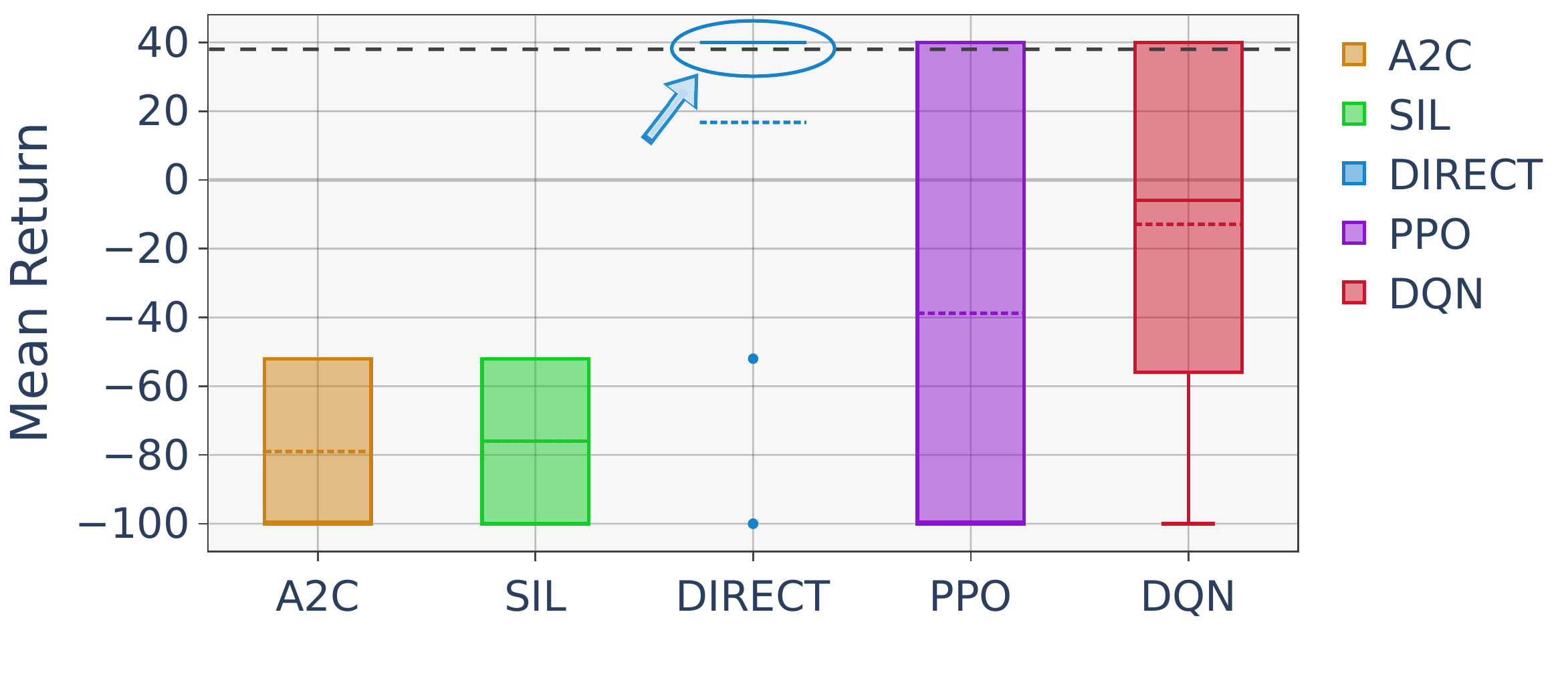}
  \caption{ Adaption Evaluation: \textmd{Episode Returns of policies trained with A2C, SIL, DIRECT, PPO, and DQN during deterministic evaluation in both the sparse and dense reward settings of the shifted target environments (cf. \autoref{fig:envs:DenseTargetShift})}}
  \label{fig:validation:trained-target}
\Description{Performance validation of trained policies in target shift environment}
\end{figure}

\section{Conclusion}
In this paper we proposed discriminative reward co-training (DIRECT) to provide intermediate feedback in challenging sparse- and shifting reward scenarios.
We first introduced the DIRECT architecture including a self-imitation buffer $\mathcal{B}$ and the corresponding update rule, a discriminator $D$ and the corresponding training loss, and an accommodating training procedure.
Our experimental results showed the directive capabilities and gained adaptability to environmental changes of DIRECT within varying safety gridworld environments.
The provided benchmark comparisons against PPO, A2C, DQN and SIL showed that DIRECT achieves superior performance especially in sparse- and shifting reward settings.

Even though outperforming state-of-the-art approaches in such challenging scenarios, DIRECT's advanced architecture, in some instances, limits fast convergence. 
Therefore, further research should consist of incorporating the discriminative reward more directly. 
Also, techniques to ensure proper convergence and reduce the risk of vanishing rewards to increase the reliability of DIRECT should be considered.
Therefore, mechanisms to steer the momentum of the self-imitation buffer would be helpful. 
Also, an increased diversity of samples within the buffer could improve the overall performance. 
Employing a diversity measure could also incentivize self conscious explorations of the environment without the need for a reward function at all \cite{eysenbach2018diversity}. 
Overall, building upon DIRECT, a broader machine learning framework based on imitation could enable generalized real-world applicability, following multiple psychological and philosophical movements that suggest that imitation is the basis for human intelligence~\cite{dennett2017bacteria}.

\balance
\bibliographystyle{bib} 
\bibliography{DIRECT.bbl}

\end{document}